# A Convolutional Neural Network for gaze preference detection: A potential tool for diagnostics of autism spectrum disorder in children

Dennis Núñez[1], Franklin Barrientos[1], Robert H. Gilman[2], Macarena Vittet[1], Patricia Sheen[1], Mirko Zimic[1*]

[1] Laboratorio de Bioinformática y Biología Molecular. Laboratorios de Investigación y Desarrollo. Facultad de Ciencias y Filosofía. Universidad Peruana Cayetano Heredia. Lima, Perú.

[2] Department of International Health, School of Public Health, Johns Hopkins University. Baltimore, USA.

*Corresponding author: Universidad Peruana Cayetano Heredia. Facultad de Ciencias y Filosofía. Laboratorios de Investigación y Desarrollo. Av. Honorio Delgado 430, SMP. Lima, Perú. Phone: (511) 3190000 ext. 2604. E-mail: mirko.zimic@upch.pe.

## Abstract

Early diagnosis of autism spectrum disorder (ASD) is known to improve the quality of life of affected individuals. However, diagnosis is often delayed even in wealthier countries including the US, largely due to the fact that gold standard diagnostic tools such as the Autism Diagnostic Observation Schedule (ADOS) and the Autism Diagnostic Interview-Revised (ADI-R) are time consuming and require expertise to administer. This trend is even more pronounced lower resources settings due to a lack of trained experts. As a result, alternative, less technical methods that leverage the unique ways in which children with ASD react to visual stimulation in a controlled environment have been developed to help facilitate early diagnosis. Previous studies have shown that, when exposed to a video that presents both social and abstract scenes side by side, a child with ASD will focus their attention towards the abstract images on the screen to a greater extent than a child without ASD. Such differential responses make it possible to implement an algorithm for the rapid diagnosis of ASD based on eye tracking against different visual stimuli.

Here we propose a convolutional neural network (CNN) algorithm for gaze prediction using images extracted from a one-minute stimulus video. The algorithm was developed from a dataset that was assembled in two stages: an initial set of adult-derived values to program the detection system and a subsequent set of child-derived values to validate the model's ability to distinguish between children with and without ASD.

In the first stage, we enrolled 40 adults between 22 and 35 years of age who worked in the Bioinformatics Laboratory of Universidad Peruana Cayetano Heredia (UPCH) in Lima, Peru. Videos were recorded in a controlled environment using a standard web camera with a frame rate of 10 images per second and with a resolution of 1280x720 pixels in RGB format.

In the second stage, we enrolled 31 children between 2 and 6 years of age who attended school in Lima, Peru. 23 of these participants had no prior history or diagnosis of ASD and attended school regularly, while 8 had a confirmed diagnosis of ASD and attended the special IMLA center (Medical Institute of Language and Learning). The diagnosis of those 8 children with ASD was confirmed by a pediatric neurologist at the IMLA special center through a combination of a clinical evaluation and previous exam findings including evocative auditive potentials, language and sensorial evaluations, and psychological evaluations.

About 30,900 images extracted from our datasets were categorized as being in one of three gaze directions and were used to train the CNN algorithm using the Caffe framework via Python and OpenCV libraries. The CNN was subsequently evaluated using cross validation techniques, testing our model's ability to predict new data that was not used in the training process, in order to flag problems like overfitting or selection bias. In this sense, our model achieved a high accuracy rate and robustness for prediction of gaze direction with independent persons and employing a different camera than the one used during testing. In addition to this, the proposed algorithm achieves a fast response time, providing a near real-time evaluation of ASD. Thereby, by applying the proposed method, we could significantly reduce the diagnosis time and facilitate the diagnosis of ASD in low resource regions.

# Introduction

Nearly 1 in 160 children worldwide are diagnosed as having autism spectrum disorder (ASD). There are two major areas associated with ASD: social communication and interaction deficits, and restrictive or repetitive behaviors. Notably, deficiencies in social reciprocity and lack of joint attention are often exhibited by children on the spectrum [10], leading to troubled behavior in school and delays in cognitive and language development [11]. These costs are borne by both individuals and their families, due to both the associated living expenses and the inherent social burden [12,13,14].

That being said, it has been shown that early intervention with children with ASD significantly improves their quality of life, with efforts aimed at modifying behavior in social interactions resulting in an improvement in the child's ability to function in such a setting [15]. Furthermore, every dollar spent on early intervention is predicted to save eight dollars in special education, crime, welfare, and other associated costs down the road [16,17].

Although certain behaviors associated with a diagnosis of ASD begin to become apparent between 18-24 months of age, the average age of diagnosis in the United States is 5.5 years [18-21]. The main reason for this trend is the suboptimal utilization of diagnostic tools such as the Autism Diagnostic Observation Schedule (ADOS) and the Autism Diagnostic Interview-Revised (ADI-R), even in higher income regions of the U.S. [22]. This is attributable to the fact that these diagnostic techniques are time-consuming (ADOS takes around 1.5 hours to administer) and require extensive training on the part of the technician (ADOS requires at least 3 days of specialized training) [23]. The ADOS technique should only be performed by certified staff, however, in developing countries there are very few trained personnel. Thus, even though ASD has been studied in-depth in low- and middle-income countries (LMICs), the true incidence and burden in many of these regions has yet to be determined.

Recent studies have shown strong evidence for gaze direction as an early biomarker of ASD [24,25,26,27]. It has been demonstrated that children with a diagnosis of ASD exhibit a preference for abstract/geometric scenes over social ones that is not present in their non-ASD counterparts [26,27]. Similarly, when looking at a person's face, children with ASD preferentially fixate on the mouth rather than the eyes when compared to children without a diagnosis of ASD [25] .

Over the past decade, classical statistical techniques have begun to be replaced by emerging Deep Learning techniques. One such approach, Convolutional Neural Networks (CNNs), has established itself as a state of the art method for object recognition, object tracking, and image segmentation [29] [30] that offers significant potential as a real-time evaluation of ASD via gaze direction. Significant computational demands and the need for sufficiently large datasets in the development of CNN-based classifiers has largely limited its use in this regard, however, although a handful of studies have reported promising results to this end. Unfortunately, the reliance of these studies on high tech cameras, simulated social interactions, and/or potentially distracting equipment limits their widespread implementation.

One such group developed a computer vision approach to analyze activities in video recordings from a study population of 6 infants, 3 of whom exhibited behaviors indicative of ASD and 3 of whom that did not [6]. They employed a dense motion estimator and a 2D body pose estimator to find differences between both groups in head motion and gait. Notably, however, the study did not attempt to classify the children based on the video recording analysis. More recently [7], eye movements were used to classify children with and without ASD. Based on the gaze patterns of all participants and using a machine learning method such as K-means clustering and a support vector machine classifier, the developed algorithm could identify children with ASD with a maximum classification accuracy of 88.51%. Nonetheless, since this process utilizes a high-acuity eye tracker, it does not lend itself for use in the large-scale screening of ASD. A third group [8] of researchers developed a screening mechanism based on a simulated social interaction. Here, each subject's voice, eye gaze, and facial expressions were tracked, and features were extracted and input into different predictive models (e.g., random forest, support vector machine, and a convolutional neural network) to detect ASD. However, since this work depends on a simulated interaction schedule with an actress, it is not well suited for younger children. One final study [9] used a CNN-based approach to predict gaze in a naturalistic social interaction to assess ASD in children. Videos of social interaction between a child and an actor were recorded by a small camera on the actor's glasses, processed, and used to develop two convolutional networks, one for face detection and another for gaze prediction. Even though gaze direction is precisely predicted by the CNNs, the use of glasses has the potential to indirectly disturb each child's attention.

While the use of gaze direction in the screening of young children for ASD holds promise, numerous challenges to its effective implementation exist. For one thing, it has been shown previously that using images obtained from a single camera can be impaired by a variety of factors including occlusions, differences in face and eye anatomies, variations of posture appearance. Moreover, many eye tracking algorithms, including open-source ones available on the internet, necessitate extensive calibration and/or restraint of the subject in order to function properly – two things that young children with ASD may be unable to endure. These algorithms can also be highly technical, require in-depth training, utilize expensive equipment, or need to be completed in highly controlled environments [26,27], limiting their practicality, especially in low-resource settings.

Here we report on a portable, non-invasive, easy-to-use, real-time, and low-cost system based on a gaze direction predictor through a CNN to measure gaze preference in children with and without ASD. We built this system using a training data set derived from images collected from adults between the ages of 24 and 35, and evaluated it on images we had previously collected of children between the ages of 2 and 5. This portable system offers promise in helping to improve the diagnosis, screening and treatment of ASD, especially in low resource regions.

In the following sections we provide a compenhise explanation of this work. In the materials and methods section we present the description of each step of the full process of gaze direction prediction as well as the explanation of the datasets and the description of the deep neural network architecture. In the results section, we present the results using metrics such as precision and confusion matrix for the different experiments performed. Finally, in the discussion section we analyze the results obtained in the previous section and describe the benefits of our work for the early detection of autism.

## Materials and Methods

**Method Overview**

The proposed research methodology follows the evidence that children with a diagnosis of ASD exhibit a preference for abstract/geometric scenes over social ones that is not present in their non-ASD counterparts [26,27]. In this sense, we use the detection of the direction of gaze towards different scenes to predict whether a given child has ASD, so this paper focuses on the use of computer vision algorithms to detect that direction of gaze. For the purposes of this work, we considered two and three gaze directions for operation on a standard personal computer. Looking at one target located on the right side of the screen (abstract scene), looking at another target located on the left side of the screen (social scene), and an additional third direction so called "undetermined", which means that the gaze is focused on a direction outside the monitor or inside the monitor but not on the established targets.

For the training data set, images were collected from adults between 24 and 35 years old who were working in the Bioinformatics laboratory of the Universidad Peruana Cayetano Heredia (UPCH). These images are formed by the adults looking in the three directions described above. In addition, for the evaluation data set we used the collection of a previous similar work developed in the Bioinformatics laboratory of the UPCH and which corresponds to children between 2 and 5 years of age taken in hospitals in the city of Lima, in these images the children appear looking at the three directions previously established.

The algorithm that we developed aims to detect the direction of the gaze within the 3 categories described as follows: the social scene, abstract scene and not determined. To accomplish this recognition task, our algorithm focuses on the eye region and more precisely on the location of the pupil within the eye region. Therefore, our first step is the extraction of these areas using a statistical cascade classifier based on local features, called Haar Cascades, which has demonstrated a high level of accuracy in detection tasks in real-time and making use of low computer resources. Once these eye regions are extracted, which are small in size compared to the total input image, they are used to train a convolutional network. Since the number of classes to be classified is three and the dimensions of the eye regions to be processed are relatively small, a LeNet-5 based architecture is chosen, which is small but with a high level of accuracy for classification tasks. Therefore, once all the images of the adult dataset have been processed, we obtain a dataset of 87,570 images with a size of 72x72 pixels. In this way, the architecture was trained with those images and testing different values for the hyper parameters of the convolutional network. After the iterative steps of variation of the hyper parameters and experimentation (training), we obtained the best values for these hyper parameters. Then, we evaluated the optimized convolutional network on the children's dataset, where we obtained promising results despite the variation in luminosity, noise, among other issues.

Our proposed system is intended to recognize the gaze direction of a child that appears on the images obtained from a video sequence. The system works as follows: First, the input frame is converted to grayscale format, next, the face region is obtained using a cascade classifier based on LBP features. Then, eyes are extracted from the facial region using a similar cascade classifier but based on Haar features, which employs Haar wavelets. Finally, the image for the CNN is generated, which contains both eye regions in a single square image. This final image is used as input for CNNf, which classifies the image into one of three classes: right direction (class 0), left direction (class 1) and not determined direction (class 2). The diagram of the proposed system is represented in Fig 1.

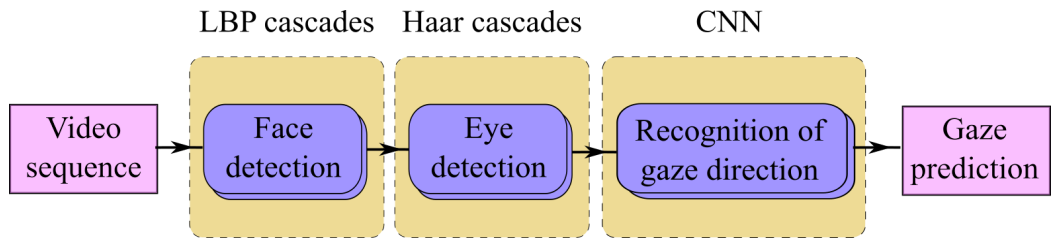

Fig 1. Diagram of the proposed system

**Study population**

*Children dataset*
For dataset collection, our team enrolled a total number of 23 children between 2-6 years old attending a regular school in Lima, Peru, with no diagnosis or previous history of ASD. In addition, our team enrolled 8 children between 2-6 years old attending the IMLA special center (Medical Institute of Language and Learning) in Lima, Peru, with a confirmed clinical diagnosis of AS.
The diagnosis of ASD for these 8 children was confirmed by a pediatric neurologist during a regular attendance of these children to the IMLA special center. The diagnosis was based on a combination of a clinical evaluation and previous exams including evocative auditive potentials, language and sensorial evaluations, and psychological evaluations. No ADOS-2 test is available currently. Although there was no quantitative score for ASD available, the clinical evaluation suggested that these children were likely high-level functioning.
This study was approved by PRISMA's Institutional Review Board (CEO921.15 April 16th, 2015). Written informed consent was obtained from parents or guardians of all children who participated in the study.

*Adult dataset*
In this dataset, a total of 40 adults between 22-35 years old, working at Bioinformatics Laboratory, at Universidad Peruana Cayetano Heredia (UPCH), Perú, were enrolled. The videos were recorded under a controlled environment and using a standard web camera with a rate of 10 frames per second with a resolution of 1280x720 pixels in RGB format.
In order to capture different eye directions, everyone was recorded looking towards one of the 7 predefined regions. The predefined regions were formed as follows: 3 areas inside the monitor screen (social, object and center) and 4 areas outside the monitor screen (up, down, right and left). Therefore, we obtained 7 videos (25 seconds each) per person, one video for each eye direction. For frame extraction, the 5 initial seconds and the 5 final seconds were not considered, so, every video has a duration of 15 seconds. Furthermore, only one frame per 3 was taken into consideration for further analysis.

**Hardware and procedure to obtain digital videos**
In order to make our database, a webcam and standard personal computer were used to capture the videos. The distance between the patient and the screen is around 1 meter and the person is looking directly at the screen. The videos were recorded in a 12 m2 environment with uniform lighting.
The screen has a LED illumination system to eliminate shadow from the person's face allowing our face detection algorithm to work without any problems. The monitor used to capture the person's gaze was 27 inches, of the brand ACER. The videos were recorded in full HD with resolution 1820 x 1200 pixels with a rate of 30 frames per second.

In the case of children, they often pay attention to their mother, and this can affect the results while we are recording the videos, so for that reason the mother has her child on her legs. However, the child sometimes wants to see his mother and they look around searching for her, and for that reason we asked the mother to hold the child's head and help the child look at the monitor. The environments were without distractions. Furthermore, during the first 10 seconds of each recording, a video of Mickey mouse was shown to the child in order to capture their attention.

**Face and eye detection**

In this study we make use of the Viola-Jones object detection framework [3], which automatically performs object detection using pre-defined features. In this way, automatic detection of faces and eyes are carried out by cascade classifiers using LBP and Haar features that were found with the Viola-Jones mechanism. Our two detectors are one after the other, which. means that first, face detection uses LBP cascades over the whole image, and then eye detection employs Haar cascades into the facial region. We make use of LBP cascades for face detection since LBP carries out calculations in integer format and Haar cascades in float format. Consequently, LBP based classifiers are faster than Haar based classifiers. Therefore, in order to find the faces over the whole image, LBP is the best option. Furthermore, since facial regions are much smaller than the whole frame, Haar cascades are a good candidate to perform an accurate eye detection. In this work we employed a robust open source face cascade detector based on the active learning described in [2].

**Gaze Direction labelling and dataset generation**

In order to distinguish three gaze directions (right, left and not determined), the children were intended to see two objects on a standard screen, one at right and another at left side of the screen. After that, we obtained 1 video per child, for a total of 27 videos of 1024x576 pixels size and one-minute each. Then, the videos were divided by 1810 frames and selected every 15 frames after the first 300 frames corresponding to the introductory attention capturing "Mickey Mouse" video. So, we generated a total of 115 frames per video. In order to predict the gaze direction using both eyes, we propose the structure depicted in Fig 2.

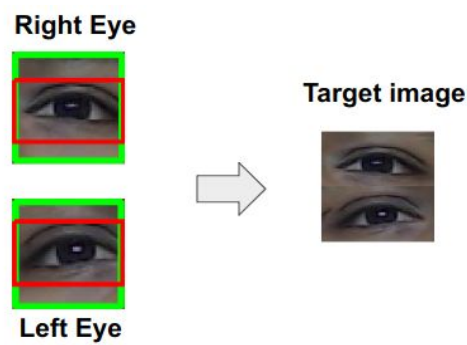

Fig 2. Composition of the image format employed to predict the gaze direction

With the purpose of having the best images for training, we cleaned the dataset by rejecting the images with poor face or eye detection. Then, manual labeling was carried out on the resulting cleaned dataset.

In the case of children, labelling was independently performed by two people. When labeling for a given image was not the same for both, it was analyzed and labeled by the two together. In this way, the labeling step was carefully inspected by a double-checking procedure.

Three gaze directions were used, each one represented by a class: right direction (class n° 0), left direction (class n° 1) and not determined direction (class n° 2). Right direction means observation of a social scene, formed by a video with a common social interaction. Left direction means observation of an abstract scene, composed by a video with abstract sequences. Not determined direction means the observation in different directions outside the monitor, which are produced by distractions in the environment or behaviors that are not related to the two main directions mentioned before.

Hence, we obtained a total of 2,060 pair-eyes images of 72x72 pixels size in RGB format. In terms of images per class, we got 704 images for class 0, 1209 images for class 1 and 147 images for class 2. Some samples of each class are depicted in Fig. 3.

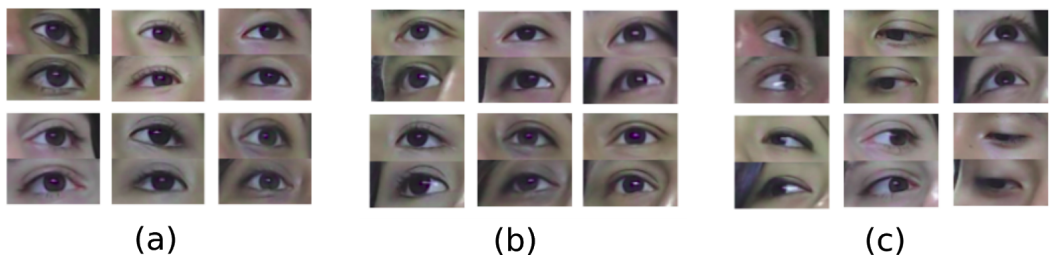

(a)            (b)            (c)

Fig 3. Samples of original RGB images obtained from children.
(a) class n° 0 or right, (b) class n° 1 or left, (c) class n° 2 or not determined

We used data augmentation to increase the training dataset. Also, to avoid overfitting and make a more robust model. Therefore, we equally shifted 5 positions along the X and Y axis by 6 pixels along such directions. Also, we employed 3 angle rotations: 10° clockwise and anti-clockwise. After that, the final data was 15 times the original, it means 30,900 images: 10,560 for class 0, 18,135 for class 1 and 2,205 for class 2. These images were used to train and test the CNN. We use grayscale images since most relevant features for eye direction prediction are color independent, and strongly rely on the shape.

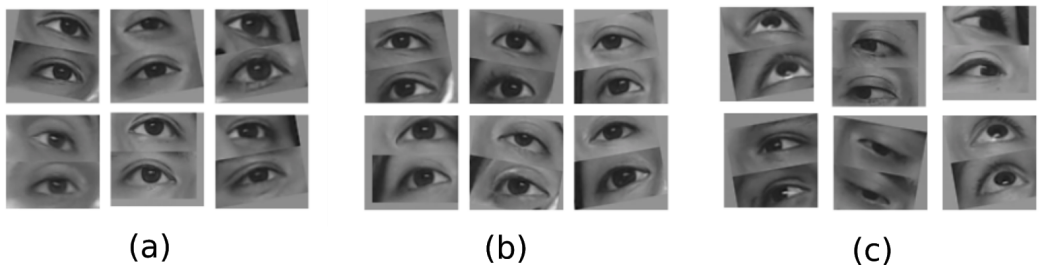

(a)            (b)            (c)

Fig 4. Samples of grayscale images obtained after data augmentation.
(a) class n° 0 or right, (b) class n° 1 or left, (c) class n° 2 or not determined

**Convolutional Neural Network for eye-tracking**

Convolutional Neural Network (CNN) is a method of Machine Learning that recently has demonstrated an outstanding performance for object recognition and image segmentation [5]. CNNs are formed by convolutional layers that work as a detection filter for localization of specific features in a training dataset and a fully connected layer that behaves as a classifier based on the non-linear combinations of the previously detected features. These features, in addition to shared weights, give the CNNs a great capacity to work with 2 dimensional arrays as inputs. Usually, CNNs are trained as typical neural networks using backpropagation.

In this way, the proposed CNN is a variation of the LeNet model [4]. So, our CNN relies on two convolutional layers, each followed by pooling operations like sub-sampling. One advantage of this sub-sampling layer is the translation invariance, which helps to position independence of our model. Furthermore, we also employed rectified linear unit (ReLU) layers, which were proposed in the Alexnet CNN [5], to add nonlinearity to the network and avoid the linearity produced by the cascades of the linear convolutions. In addition to this, the proposed model makes use of dropout regularization to ignore neurons during training, avoiding overfitting of the model. Finally, the second convolutional layer is followed by a fully connected (FC) layer of 120 neurons and a 3-way SoftMax layer, (Fig. 5). The combination of the layers mentioned before gives a total of 60K learnable parameters.

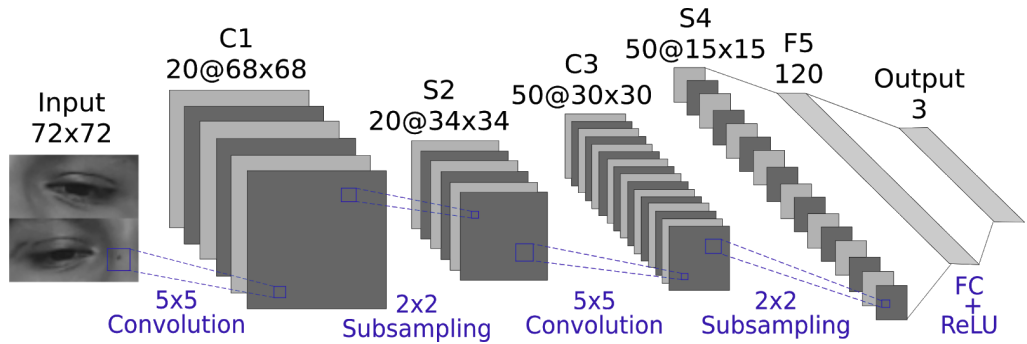

Fig 5. Proposed CNN architecture.

**CNN performance analysis**
In order to evaluate the correct performance of our CNN model, it was evaluated in two different ways: 5-fold cross-validation and independent camera - independent dataset.

*5-fold Cross-Validation*
For better statistics of our model and to prevent overfitting, we evaluated our model using a 5-fold cross-validation and employing groups of independent people that do not appear in the training dataset. In this last way of evaluating our model rigorously testing is carried out in a different dataset.

*Using 3 classes*
The whole dataset is divided into 5 nearly equally sized groups: 1st group with 17,895 images, 2nd group with 17,670 images, 3rd group with 17,700 images, 4th group with 17,085 images and 5th group with 17,220 images.
For all iterations, the training step was carried out by 6,000 iterations with a batch of 100.

*Using 2 classes*
The whole dataset is divided into 5 nearly equally sized groups: 1st group with 11,820 images, 2nd group with 11,775 images, 3rd group with 11,490 images, 4th group with 11,470 images and 5th group with 11,025 images.
For all iterations, the training step was carried out by 6,000 iterations with a batch of 100.

**Testing on children's dataset (Independent Person - Independent Dataset)**
The evaluation of the proposed CNN using an independent camera, which was not used to capture the training dataset, and one person that does not appear in the training dataset. The camera used was a laptop webcam, which generated a video of 1280x720 pixels of a one-minute duration. Training was carried out by 6,000 iterations with a batch of 100.

# Results

## CNN performance analysis

### 3-classes
The results of each fold are shown in Table 1. Based on this table, for the 5-fold cross-validation, we calculate an average accuracy of 89.54%.

|  | **Fold 1** | **Fold 2** | **Fold 3** | **Fold 4** | **Fold 5** |
|---|---|---|---|---|---|
| **Accuracy** | 89.65 | 88.61 | 90.45 | 93.29 | 85.70 |

Table 1. Results of 5-fold cross-validation (%)

### 2-classes
The results of each fold are shown in Table 2. Based on this table, for the 5-fold cross-validation, we calculate an average accuracy of 97.38%.

|  | **Fold 1** | **Fold 2** | **Fold 3** | **Fold 4** | **Fold 5** |
|---|---|---|---|---|---|
| **Accuracy** | 97.81 | 96.48 | 98.40 | 98.64 | 95.57 |

Table 2. Results of 5-fold cross-validation (%)

## Testing on children's dataset (Independent Person - Independent Dataset)

### For 3 classes
The results show an accuracy of 88.57%, confusion matrix is shown in Fig 9.

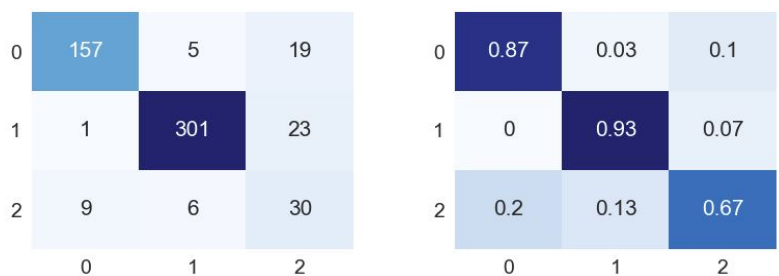

Fig. 6. Confusion matrix with normalized values

**For 2 classes**
The results show an accuracy of 95.12%, confusion matrix is shown in Fig 7.

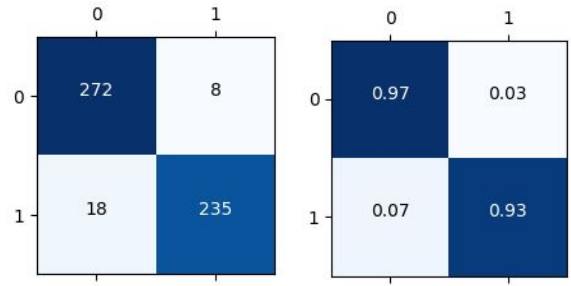

Fig. 7. Confusion matrix with normalized values

# Discussion

This study shows that our proposed method achieved a high classification accuracy of 95.1%. The testing step was performed on a different dataset than the training one, which was formed by the children's dataset. In addition, it should be noted that the testing dataset was obtained with a different camera than the one used to capture the training images. These considerations of independence of the datasets and different cameras to perform the evaluation of our model makes this evaluation very rigorous but at the same time verify its performance in a better way for the purposes of this project.

The main advantage of our algorithm is that it makes use of convolutional networks, which have shown an advantage in different classification, segmentation and detection tasks compared to classical image processing algorithms. Indeed, convolutional neural networks have become the state of the art for object recognition in computer vision [5]. During the last years, CNNs have been applied for classification and detection tasks on medical images, including classification of 2D HeLa cells in fluorescence microscopy [31], classification of Apoptosis in phase contrast microscopy [32], and classification of breast tissue [33].

In a previous study [28], the initial videos were recorded in a non-optimal way, so several problems occurred. Some of these problems were as follows: poor lighting, disturbance of the eyelashes from some angles, small screen (did not allow to discrimination of right/left glance direction), the supervisor did not hold the child's head (this produced un-processable images due to movements of the child's head), the camera didn't have autofocus and a poor resolution (which results in low eye region quality). Therefore, the results of deep learning techniques on the final recorded videos, which were recorded in optimal conditions (mentioned in subsection 2.3), were by far better than the results on the initial non- optimal videos. As for the experimental results for a different number of classes, when we consider 3 gaze directions (3 classes), misclassification errors occur mainly due to undetermined direction (class n° 2). Therefore, adult glance direction at near-the-border positions produces confusion with right and left gaze directions, class number 0 and class number 1 respectively. Evidence of this is shown in matrix confusion, see Fig. 6.

In addition to this published work, several popular open-source frameworks for eye tracking are available on the web. Unfortunately, not all are well suited for autism screening. For instance, one of the most popular, Pupil [https://github.com/pupil-labs/pupil], also makes use of glasses connected to a computer device, so it, too, could disrupt a child's concentration and result in an incorrect diagnosis of autism. Similarly, another tool, PyGaze [http://www.pygaze.org], specialized cameras that are expensive and difficult to acquire, limiting the framework's practicality. While this latter program can be used with common webcams, as well, this poses challenges for calibration due to the need for a child's eyes to remain in a fixed position. Finally, other open source tools like WebGazer [https://webgazer.cs.brown.edu/] and EyeTracker [https://github.com/hugochan/Eye-Tracker], are limited for use with web browsers and smartphones, respectively.

The development of a gaze detection system using Deep Learning techniques would have great benefits for the early diagnosis of Autism, given that there is a strong need for a fast, effective and accessible methodologies for early autism diagnosis in children, especially in developing countries where gold standard diagnostic tools such as the Autism Diagnostic Observation Schedule (ADOS) and the Autism Diagnostic Interview-Revised (ADI-R) are difficult to implement due to their requirement for certified personal and their consuming nature. In addition, recent studies have shown that early intervention in children with ASD significantly improves their quality of life. Activities that involve behavioral modifications in social interactions between children result in improved individual capacity in social settings [15]. In addition, every dollar spent on early intervention helps families save eight dollars in special education, crime, welfare and other associated costs [16,17]. Despite the high importance of accessible methodologies for early autism diagnosis and the different approaches using Computer Vision techniques, only a few articles demonstrated complete solutions where the images are taken by a standard RGB web camera and both detection and classification algorithms can demonstrate promising results in real-time.

In conclusion, our results show that our proposed method achieves a high classification accuracy of 95.1% for a rigorous test, which was carried out on a different dataset (children dataset) recorded with a different camera. Furthermore, the response time at inference step shows a real-time response of about 20 ms. In this sense, the present work shows the design of an efficient gaze tracking system using a convolutional neural network. At the same time, this system shows advantages in comparison to previous works in terms of better classification accuracy, greater robustness to noise and better response time. Therefore, the previous results demonstrate that the proposed system is useful for a better, effective and accessible autism diagnosis tool.